\documentclass[letterpaper, 10 pt, conference]{ieeeconf}  %

\IEEEoverridecommandlockouts                              %

\usepackage{adjustbox}
\usepackage{url}            %
\usepackage{xfrac}       %
\usepackage{booktabs}       %
\usepackage{microtype}      %
\usepackage{xcolor}         %
\usepackage{textcomp}

\usepackage{amsmath}
\usepackage{amssymb}
\usepackage{amsfonts}
\usepackage{bm}
\usepackage{mathtools}
\usepackage{cancel}

\usepackage{enumitem}
\usepackage{pifont}

\usepackage{listings}
\definecolor{codegreen}{HTML}{85AB59}
\definecolor{codegray}{rgb}{0.5,0.5,0.5}
\definecolor{codepurple}{rgb}{0.58,0,0.82}
\definecolor{backcolour}{HTML}{F5F5F5}
\definecolor{codeblue}{HTML}{2B5BB5}
\definecolor{codecyan}{HTML}{2BBAC5}

\lstdefinestyle{mystyle}{
    language=Python,
    backgroundcolor=\color{backcolour},   
    commentstyle=\color{green},
    keywordstyle=\color{magenta},
    numberstyle=\tiny\color{codegray},
    stringstyle=\color{codegreen},
    basicstyle=\ttfamily\footnotesize,
    columns=fullflexible,
    breakatwhitespace=false,         
    breaklines=true,                 
    captionpos=b,
    xleftmargin=1em,
    keepspaces=true,                 
    numbers=left,                    
    numbersep=1em,                  
    showspaces=false,                
    showstringspaces=false,
    showtabs=false,                  
    tabsize=2,
    escapechar={|}
}

\usepackage{wrapfig}
\usepackage{float}
\usepackage{graphicx}
\usepackage[caption=false,font=footnotesize]{subfig}

\usepackage[ruled,vlined,linesnumbered]{algorithm2e}
\usepackage{algcompatible}
\usepackage{multirow}

\usepackage[acronym]{glossaries}

\usepackage{cite}

\definecolor{cvprblue}{rgb}{0.21,0.49,0.74}
\definecolor{ieeeblue}{HTML}{035B98}
\usepackage[breaklinks,colorlinks,linkcolor=ieeeblue,citecolor=ieeeblue,urlcolor=ieeeblue]{hyperref}

\usepackage{orcidlink}

\newacronym{GNN}{GNN}{graph neural network}
\newacronym{CNN}{CNN}{convolutional neural network}
\newacronym{MDN}{MDN}{mixture density network}
\newacronym{CVAE}{CVAE}{conditional variational autoencoder}
\newacronym{GAN}{GAN}{generative adversarial network}
\newacronym{WOMD}{\emph{WOMD}}{Waymo Open Motion Dataset}
\newacronym{BEV}{BEV}{bird's-eye view}
\newacronym{POV}{POV}{point of view}

\newacronym{MHA}{MHA}{multi-head attention}
\newacronym{GGRU}{Graph-GRU}{graph-gated recurrent unit}
\newacronym{GRU}{GRU}{gated recurrent unit}

\newacronym{HD}{HD}{high-definition}

\newacronym{ADE}{ADE}{Average Displacement Error}
\newacronym{FDE}{FDE}{Final Displacement Error}
\newacronym{BFDE}{Brier-FDE}{Brier Final Displacement Error}
\newacronym{MR}{MR}{Miss Rate}
\newacronym{CR}{CR}{Collision Rate}
\newacronym{APDE}{APDE}{Average Path Displacement Error}
\newacronym{ANLL}{ANLL}{Average Negative Log-Likelihood}
\newacronym{FNLL}{FNLL}{Final Negative Log-Likelihood}

\newacronym{GCN}{GCN}{graph convolutional network}
\newacronym{GAT}{GAT}{graph attention network}

\newacronym{TA}{TA}{target agent}
\newacronym{TTLC}{TTLC}{time-to-lane-change}

\newacronym{MA}{MA}{Multi-agent}
\newacronym{SA}{SA}{Single-agent}

\newacronym{PYG}{PyG}{PyTorch Geometric}
\newacronym{UAV}{UAV}{unmanned aerial vehicle}

\DeclareMathOperator*{\argmin}{argmin}

\newcommand{\R}{\mathbb{R}}

\newcommand\norm[1]{\lVert#1\rVert}

\definecolor{MyGreen}{HTML}{10A23B}
\definecolor{MyRed}{HTML}{DD1111}

\newcommand{\node}{\mathcal{V}}

\newcommand{\stateestim}[1]{\hat{\bm{x}}_{#1}}
\newcommand{\mix}[1]{\bm{\pi}^{#1}}

\newcommand{\state}[1]{\bm{x}_{#1}}

\newcommand{\predhrz}{N}

\usepackage{abbrv}

\title{\LARGE \bf
	Toward Unified Practices in Trajectory Prediction Research on \\ Bird's-Eye-View Datasets
}

\author{Theodor Westny$^{1}$\orcidlink{0000-0001-9075-7477}, Bj\"orn Olofsson$^{2,1}$\orcidlink{0000-0003-1320-032X}, and Erik Frisk$^{1}$\orcidlink{0000-0001-7349-1937}%
\thanks{This research was supported by the Strategic Research Area at Linköping-Lund
in Information Technology (ELLIIT) and the Wallenberg AI, Autonomous Systems and Software Program (WASP) funded by the Knut and Alice Wallenberg Foundation.}%
\thanks{$^{1}$Department of Electrical Engineering,
 	Linköping University, Sweden.}%
\thanks{$^{2}$Department of Automatic Control,
 	Lund University, Sweden.}%
\thanks{Corresponding author e-mail: {\tt\footnotesize theodor.westny@liu.se}}%
}

\begin{document}

\maketitle
\thispagestyle{empty}
\pagestyle{empty}

\begin{abstract}
    The availability of high-quality datasets is crucial for developing behavior prediction algorithms in autonomous vehicles.
    This paper highlights the need to standardize the use of certain datasets for motion forecasting research to simplify comparative analysis and proposes a set of tools and practices to achieve this.
    Drawing on extensive experience and a comprehensive review of current literature, we summarize our proposals for preprocessing, visualization, and evaluation in the form of an open-sourced toolbox designed for researchers working on trajectory prediction problems.
    The clear specification of necessary preprocessing steps and evaluation metrics is intended to alleviate development efforts and facilitate the comparison of results across different studies.
    The toolbox is available at: {\footnotesize\url{https://github.com/westny/dronalize}}.
\end{abstract}

\section{Introduction}
\label{sec:introduction}
The trajectory-prediction problem has seen a surge in interest over the past decade~\cite{huang2022survey, fang2024behavioral}.
The increased research activity has been encouraged by the development of autonomous vehicles, which requires methods that can understand and anticipate the future behavior of other road users.
As the field has moved toward more data-driven approaches~\cite{huang2022survey, fang2024behavioral}, the availability of high-quality datasets has become crucial for the development of trajectory-prediction algorithms.
This also emphasizes the need for standardized benchmarking suites that allow researchers to evaluate the performance of their methods and compare their results with other studies.

For a long time, researchers have relied on traffic datasets such as the I-80, US-101, Lankershim, and Peachtree datasets from the Next Generation Simulation (NGSIM) program~\cite{alexiadis2004next} that have been widely used for studies on behavior prediction problems~\cite{phillips2017generalizable,deo2018convolutional,hu2018probabilistic,zhao2019multi,ding2019predicting,mercat2020multi,westny2021vehicle}.
The NGSIM datasets include data from the perspective of a stationary camera overlooking a region of interest, providing a \gls{BEV} of the traffic scene.
However, some of these datasets suffer from high noise levels and tracking errors, which makes them less reliable for traffic studies~\cite{montanino2013making, coifman2017critical}. 
In response, recent years have seen the release of new datasets that aim to address some of these issues, such as \emph{SDD}~\cite{robicquet2016learning}, \interact~\cite{zhan2019interaction}, \sind~\cite{sinDdataset} and those from \emph{leveLXData}~\cite{highDdataset,rounDdataset,inDdataset,exiDdataset,uniDdataset}.
These datasets are primarily collected using camera-equipped aerial drones and provide data of particularly high spatial and temporal resolution.
However, a challenge in using some of these datasets is the lack of a widely recognized standard for preprocessing steps---making it difficult to reproduce results and compare performances across various trajectory prediction studies.

In contrast to the aforementioned datasets, there are also those collected by intelligent vehicles, \eg, \emph{HDD}~\cite{ramanishka2018toward}, \emph{NuScenes}~\cite{caesar2020nuscenes}, \emph{Argoverse~2}~\cite{argoverse2dataset}, and \gls{WOMD}~\cite{ettinger2021large}, which consist of data from onboard sensors, and thus from the \gls{POV} of the vehicle.
These datasets have seen widespread adoption in the research community, in particular \argoverse{} and \gls{WOMD}, as they provide benchmarking suites for trajectory prediction problems and appear in workshop challenges~\cite{wadWorkshopAutonomous}.
Because the data are collected using the actual sensor suite of the vehicle, their use facilitates model development adopted for real-world conditions.
Despite their realism, however, these datasets may not always capture all surrounding objects that influence the behavior of other agents because of potential sensor occlusions.
This aspect introduces uncertainty in evaluations, making it challenging to discern the specific contributions of the model and the data in highly interactive scenarios.
Importantly, we argue that neither \gls{BEV} nor \gls{POV} type of datasets are preferable over the other; rather, they should be complementary in trajectory prediction studies.
Inspired by the most recent \gls{POV}-style datasets, we introduce a standardized preprocessing pipeline and evaluation protocol for trajectory prediction research on several \gls{BEV} datasets; in particular, \highd~\cite{highDdataset}, \round~\cite{rounDdataset}, \ind~\cite{inDdataset}, \exid~\cite{exiDdataset}, \unid~\cite{uniDdataset}, \sind~\cite{sinDdataset}, \isac~\cite{isacdataset}, and \interact~\cite{zhan2019interaction}.
By reconciling various approaches from the existing literature, we aim to establish a standard that reflects the most common practices in the field.

\subsection{Contributions}
\label{sec:contributions}
The contribution of this paper is the proposal of standardized practices for trajectory prediction research on \gls{BEV} datasets. %
These practices are realized in the form of a toolbox that serves two primary purposes:
\begin{itemize}
    \item By standardizing preprocessing procedures, researchers can focus more on algorithm development rather than data preparation.
    \item A clear specification of evaluation metrics, which, when combined with standardized preprocessing, facilitates direct comparisons across studies without the need for re-implementing existing methods.
\end{itemize}
When combined, these properties help to alleviate the development efforts of research pipelines, thereby accelerating the discoveries in the trajectory-prediction field.
Implementations were made in PyTorch~\cite{paszke2019pytorch}. %
The toolbox is available at: {\small\url{https://github.com/westny/dronalize}}.

\section{Background}
\label{sec:background}
The focus on the aforementioned datasets is driven by several factors.
First, their high tracking accuracy allows for interpretable and reliable evaluations---making them particularly suitable for early-stage research and development.
Second, these datasets feature a diverse range of interacting road users, including cars, trucks, and vulnerable road users (VRUs) such as pedestrians and cyclists, providing rich scenarios for studying safety-critical interactions in autonomous driving applications.
The data were collected at several fixed locations (highways, intersections, roundabouts, \etc) across Germany, China, and the United States, providing a broad range of traffic scenarios.
The datasets are made up of multiple recordings that contain both static and dynamic information, provided in {\small\texttt{.csv}} files. 
All but \highd{} are also accompanied by a Lanelet~2~\cite{poggenhans2018lanelet2} file in {\small\texttt{.osm}} format, which can be used to extract semantic information about the scene.

\section{Related Work}
Although there are numerous examples of studies that utilize the aforementioned datasets for behavior prediction research~\cite{diehl2019graph,shao2023self,shao2023failure,wei2024ki,ma2021continual,carrasco2021scout,cheng2021exploring,wen2022social,messaoud2020attention,chen2022intention,westny2023mtp,westny2023eval,mozaffari2023multimodal,mozaffari2023trajectory,wang2024spatio}, they are used in different ways, some of which are illustrated in Table~\ref{tab:study_specs}.
These inconsistencies complicate the comparison of results across studies, as fair evaluation requires uniform conditions for all methods. 
Rather than advocating for one approach over another, we aim to propose a standard that reflects the most common practices in the field.
Related efforts include tools that provide unified interfaces to facilitate data access and evaluation across well-curated trajectory prediction datasets, such as \cite{ivanovic2023trajdata} and \cite{feng2024unitraj}.
In contrast, our focus is on streamlining benchmarking practices for frequently used BEV datasets that \emph{lack} established preprocessing and evaluation protocols.
As such, our work is closely aligned with initiatives that define clear dataset partitions and evaluation metrics in the context of structured challenges such as \cite{caesar2020nuscenes}, \cite{argoverse2dataset}, and \cite{ettinger2021large}.

\begin{table}[!h]
	\caption{Time specifications of studies using BEV datasets.}
	\label{tab:study_specs}
	\centering
    \resizebox{\columnwidth}{!}{%
       {
    \setlength{\tabcolsep}{8pt}
	\begin{tabular}{c c c c}
		\toprule
		Study & Sampl. Time & Observ. Window & Pred. Horizon\\
		\midrule
        \cite{diehl2019graph} & $1.0$~s & $5.0$~s & $5.0$~s\\
        \cite{shao2023self, shao2023failure} & $0.5$~s & $3.0$~s & $3.0$~s\\
        \cite{wei2024ki} & $0.5$~s & $6.0$~s & $6.0$~s\\
        \cite{ma2021continual} & $0.4$~s & $1.2$~s & $3.2$~s\\
        \cite{carrasco2021scout,cheng2021exploring} & $0.4$~s  & $3.2$~s & $4.8$~s \\
        \cite{wen2022social} & $0.4$~s & $4.0$~s & $4.0$~s \\
        \cite{messaoud2020attention,chen2022intention,westny2023mtp,westny2023eval,mozaffari2023multimodal,mozaffari2023trajectory} & $\bm{0.2}$~s  & $\bm{3.0}$~s & $\bm{5.0}$~s \\
        \cite{wang2024spatio} & $0.08$~s & $1.6$~s & $2.4$~s \\
		\bottomrule
	\end{tabular}
       }
    }
\end{table}

\section{Preprocessing}
\label{sec:preprocessing}

\begin{figure}[!t]
    \centering
    \includegraphics[width=0.9\columnwidth]{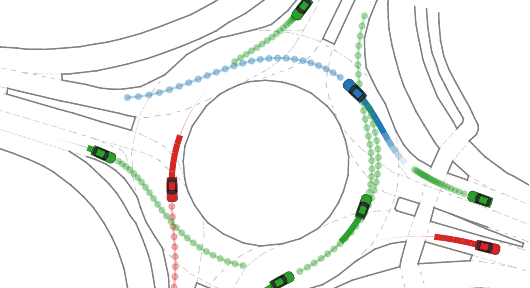}
    \caption{Example scenario from the \round{} dataset.
    The color of the agents represents how they are scored in the prediction task.
    The blue vehicle is the single-agent target, the green vehicles are the multi-agent targets, and the red vehicles are the non-scored surrounding agents.
    The same-colored lines following each agent represent the observed past trajectory, while the same-colored dotted lines preceding them indicate their future trajectory.
    }
    \label{fig:scenario-ex}
\end{figure}

\subsection{Prediction Objective}
\label{sec:prediction-objective}
Resources for both single-agent and multi-agent trajectory prediction tasks are provided in the toolbox.
In all cases, information on surrounding agents is supplied (should they exist) to encourage interaction-aware mechanisms.
Motivated by the literature, see Table~\ref{tab:study_specs}, the observation window (length of the input sequence) is set to a maximum of $3$~s, while the prediction horizon is set to a maximum of $5$~s.
Each prediction scenario is centered around a \gls{TA}, which also functions as the objective for the single-agent prediction task.
The information on the single-agent target is always constructed to have maximum-length ground-truth data.
To ensure consistency across scenarios, up to $8$ nearest neighbors to the \gls{TA} in the last observed frame are selected for the multi-agent prediction task---provided they have at least $3$~s of future ground-truth data available.
An example scenario is shown in Fig.~\ref{fig:scenario-ex}.

\begin{figure}[!t]
    \centering
    \includegraphics[width=0.96\columnwidth]{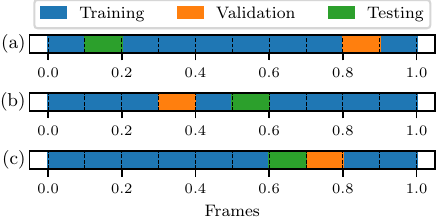}
    \caption{Example of three different data splits, denoted by (a), (b), and (c), using the proposed partitioning method.
    The figure illustrates how the bins are divided into training, validation, and test sets based on the number of frames in each recording.
    The different colors are used to indicate which set each bin belongs to, where the dashed lines represent the boundaries between the bins. 
    }
    \label{fig:datasplit}
\end{figure}

\subsection{Dataset Splits}
\label{sec:dataset-splits}
Dataset partitioning is conducted at the recording level based on the number of frames in each file.
For every recording, the total number of frames is partitioned into $10$ equally sized bins, which are then randomly assigned to the training, validation, and test sets using a rough $80\%$-$10\%$-$10\%$ split, as illustrated in Fig.~\ref{fig:datasplit}.
Since traffic behavior is highly dependent on the time of day,
it is important to ensure that all partitions contain segments from various frames throughout the recordings to mitigate selection bias.
The data extraction process then proceeds by collecting trajectories to construct scenarios from the recordings in the allocated time frames.
Overlapping scenarios between temporally adjacent bins within the same partition are permitted to maximize the collection of samples. To avoid data leakage between the training, validation, and test sets, it is ensured that no agent is present in more than one partition.
Using the proposed preprocessing approach, the number of scenarios for each partition and dataset is summarized in Table~\ref{tab:num_samples}, with the total number of trajectories shown in parentheses.
We note that the \interact{} dataset has predefined splits, which we follow in our preprocessing pipeline.
\begin{table}[!t]
	\caption{Number of scenarios per dataset and partitions.}
	\label{tab:num_samples}
	\centering
    \resizebox{\columnwidth}{!}{%
        {
    \setlength{\tabcolsep}{6pt}
	\begin{tabular}{l l l l}
		\toprule
		Dataset & Training & Validation & Testing\\
		\midrule
		\highd & 103,489 (1,532,748)  & 12,549 (184,787) & 13,748 (202,499) \\
        \round & 162,071 (1,793,089) & 18,811 (209,376) & 18,599 (204,420) \\
        \ind & 103,587 (2,227,838) & 12,232 (260,246) & 11,781 (254,255) \\
		\unid & 124,456 (3,092,743) & 15,758 (392,394) & 15,792 (436,126) \\
		\exid & 252,907 (3,054,668) & 30,635 (379,079) & 29,344 (361,291) \\
		\sind & 199,381 (5,730,586) & 26,046 (814,105) & 21,657 (560,941) \\
		\isac & 28,031 (1,085,116) & 4,766 (234,932) & 5,600 (173,219) \\
		\interact & 47,584 (537,599) & 11,794 (141,395) & 2,644 (30,659) \\
		\bottomrule
	\end{tabular}
        }
    }
\end{table}

It is important to recognize that there is a notable imbalance across the maneuver classes within the highway datasets (the lane-change percentage is around $13\%$).
To ensure that the scenarios are both dynamic and challenging, we adopt a stratified sampling approach, aiming to achieve a more balanced distribution of the \gls{TA} maneuvers---the results of which are shown in Table~\ref{tab:man_dist}.
We use this information to label the trajectories with the corresponding maneuver class, which can be used for assessing the ability of the model to predict specific maneuvers as well as intention prediction applications.
Seven maneuver classes are employed based on the \gls{TTLC}, as described in Fig.~\ref{fig:maneuver_classes}.
\begin{table}[!t]
	\caption{Number of TA highway maneuvers per dataset.}
	\label{tab:man_dist}
	\centering
    \resizebox{\columnwidth}{!}{%
        {
    \setlength{\tabcolsep}{10pt}
	\begin{tabular}{l l l l}
		\toprule
		Dataset & Lane change left & Lane keeping & Lane change right\\
		\midrule
		\highd & 30,634 (23,6\%) & 63,189 (48,7\%) & 35,963 (27,7\%) \\
		\exid & 106,490 (34,0\%) & 147,347 (47,1\%) & 59,029 (18,9\%) \\
		\isac & 6,090 (15,9\%) & 21,654 (56,4\%) & 10,653 (27,7\%) \\
		\bottomrule
	\end{tabular}
        }
    }
\end{table}

\begin{figure}[h]
    \centering
    \resizebox{\columnwidth}{!}{%
    \begin{tikzpicture}[
        level 1/.style={sibling distance=5cm},
        level 2/.style={sibling distance=2cm},
        every node/.style={draw,rounded corners,align=center},
    ]
        \node {Maneuver Classes}
            child {node {Lane Change\\ Left (LCL)}
                child {node {0:\\ TTLC $\leq$ 1s}}
                child {node {1:\\ TTLC $\leq$ 3s}}
                child {node {2:\\ TTLC $\leq$ 5s}}
            }
            child {node {Lane\\ Keeping (LK)}
                child {node {3:\\ Maintaining\\ current lane}}
            }
            child {node {Lane Change\\ Right (LCR)}
                child {node {4:\\ TTLC $\leq$ 1s}}
                child {node {5:\\ TTLC $\leq$ 3s}}
                child {node {6:\\ TTLC $\leq$ 5s}}
            };
    \end{tikzpicture}
    }
    \caption{Maneuver class hierarchy and corresponding values.}
    \label{fig:maneuver_classes}
\end{figure}

\subsection{Coordinate System}
\label{sec:coordinate-system}
Although the datasets are consistent in terms of the data they provide, some differences need to be addressed, such as the coordinate system used.
This is mainly an issue when working with the \highd{} dataset.
To ensure that the samples are consistent across the datasets, the data are first divided into two separate groups based on the direction of travel.
Next, the data are transformed to a common coordinate system where the $x$-axis points in the direction of the road, the $y$-axis points to the left, and the origin is moved to the lower left corner of the current road segment.
For the other datasets, there are no such obvious issues, as all agents move in the same area. 
Regardless, we still place the origin in the approximate geographical center of the scene to ensure some consistency across recordings.

A potential issue with using a global coordinate system is that it may introduce prediction bias, where the model learns scene-specific patterns rather than generalizable behavior---an essential property for zero-shot learning~\cite{jaipuria2018curbside,hu2022scenario}.
While we leave this design choice to the user, one possible approach is to transform the scenario into a relative (agent-centric) coordinate system, such as the local frame of the \gls{TA}.

\subsection{Downsampling}
\label{sec:downsampling}
\begin{figure}[!t]
    \centering
    \includegraphics[width=0.85\columnwidth]{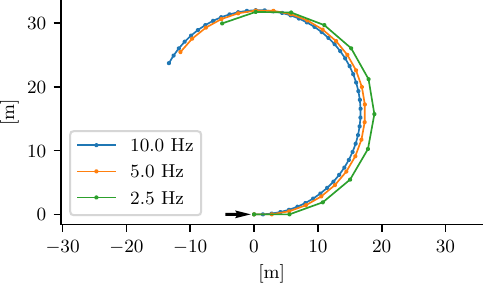}
    \caption{Example simulation of a vehicle moving along a circular path with a radius of $16$~m, a speed of $50$~km/h (speed limit) for a duration of $5$~s using different sampling rates in a forward Euler integration scheme.
    The values represent nominal conditions for the \round{} dataset.
    The figure is used to illustrate how different sampling rates can affect motion prediction quality.
    This is something that could be of interest when assessing the physical feasibility of the predicted trajectories.
    }
    \label{fig:roundabout-ex}
\end{figure}

Most related works have used downsampling as a preprocessing step~\cite{diehl2019graph,messaoud2020attention,carrasco2021scout,ma2021continual,cheng2021exploring,quintanar2021predicting,chen2022intention,wen2022social,westny2023mtp,westny2023eval,mozaffari2023multimodal,wang2024spatio}, albeit with different integer factors.
Although there are several potential computational benefits---allowing for faster prototyping and experimentation---it is important to consider the qualitative effect downsampling has on both the data and the model.
Figure~\ref{fig:roundabout-ex} is used to illustrate how different sampling rates can affect the quality of predicted trajectories using nominal driving conditions for the \round{} dataset using a 1\textsuperscript{st}-order method.
In the figure, we see that for lower sampling rates, the simulated trajectories become overly jagged and unrealistic, which makes it difficult to assess the physical feasibility of the predicted trajectories.
To ensure that the data are still representative of the original tracking data, we downsample it to a rate of $5$~Hz, corresponding to a sampling time of $0.2$~s.
When studying the frequency spectrum of the tracking data, it was found that several signals have components above the new Nyquist frequency ($2.5$~Hz). %
Before downsampling, an anti-aliasing filter is applied to remove these high-frequency components using a 7\textsuperscript{th} order Chebyshev type I filter (based on qualitative evaluations).

\subsection{Agent Features}
\label{sec:agent-features}

\subsubsection{Tracking Features}
The datasets contain a rich set of features for each agent, which are mostly consistent across the datasets.
In Table~\ref{tab:agent_feat} we list the type of features that are made available for each agent present in a sample scenario.
The types of features available are the same in both the input and ground-truth target data. %
Importantly, the heading angle is not available in the \highd{} dataset, and is therefore estimated using the instantaneous velocities as 
\begin{equation}
    \psi = \arctan2(v_y, v_x).
\end{equation}
Longitudinal and lateral accelerations are included as supplementary features. However, we discourage their direct use in the final model, as such measurements of high quality are rarely available in \gls{POV} data. Estimated values (\eg, via finite differences) may still be used when appropriate.

\begin{table}[!t]
	\caption{Agent features.}
	\label{tab:agent_feat}
	\centering
    \resizebox{\columnwidth}{!}{%
        {
    \setlength{\tabcolsep}{10pt}
	\begin{tabular}{l l l}
		\toprule
		Feature & Description & Unit\\
		\midrule
		$x$ & Local x coordinate& m\\
		$y$ & Local y coordinate & m\\
		$v_x$ & Velocity in the x-axis direction & m/s \\
		$v_y$ & Velocity in the y-axis direction & m/s\\
		$\psi$ & Heading angle  & rad \\
		\multicolumn{3}{c}{\emph{Supplementary}}\\
		\midrule
		$a_x$ & Acceleration in the x-axis direction & m/$\text{s}^2$ \\
		$a_y$ & Acceleration in the y-axis direction & m/$\text{s}^2$\\
		\bottomrule
	\end{tabular}
        }
    }
\end{table}

\subsubsection{Agent Classes}
\label{sec:agent-classes}
The different datasets contain agents of different types, \eg, cars, bicycles, pedestrians, \etc.
We hypothesize that the model could benefit from the inclusion of this information, as different classes exhibit different behaviors.
To allow for a more generalizable model, we propose a common set of agent classes across the datasets.
The categories could be included as one-hot encoded features, or directly used as tokens in an embedding layer. 
The distribution of unique agents and their respective classes in the datasets is shown in Table~\ref{tab:class_dist}.
We note that the number of unique agents in the \interact{} dataset is not explicitly provided, although the accompanying paper reports a total of 40,054 recorded vehicles.
For reference, our preprocessed data contains 676,931 vehicle trajectories, along with 32,722 trajectories corresponding to a single VRU class.

\begin{table}[!h]
	\caption{Number of unique agent classes per dataset.} 
	\label{tab:class_dist}
	\centering
    \resizebox{\columnwidth}{!}{%
        {
    \setlength{\tabcolsep}{6pt}
	\begin{tabular}{l l l l l l l l}
		\toprule
		Dataset & Car & Truck & Bus & Motorcycle & Bicycle & Pedestrian & Tricycle \\
		\midrule
		\highd & 79,757 & 19,722 & --- & --- & --- & --- & --- \\
        \round & 12,103 & 1,047 & 51 & 121 & 87 & 23 & ---  \\
        \ind & 7,402 & --- & 330 & --- & 2,239 & 3,102 & --- \\
		\unid & 1,302 & 38 & 28 & 36 & 706 & 7,835 & ---  \\
		\exid & 54,485 & 12,123 & --- & 74 & --- & --- & --- \\
		\sind & 20,574 & 823 & 103 & 4,660 & 3,267 & 4,159 & 729  \\
		\isac & 7,138 & 864 & 170 & 445 & --- & --- & ---  \\
		\interact & $\approx$ 40,000 & --- & --- & --- & Unknown & Unknown & ---  \\
		\bottomrule
	\end{tabular}
        }
    }
\end{table}

\begin{figure*}[!t]
	\centering
	\subfloat[Lane graph based on a Lanelet from the \round{} dataset.]{\includegraphics[width=0.5\textwidth]{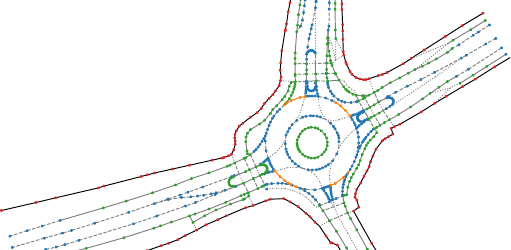}%
        \label{fig_first_case}}
	\hfill
    \subfloat[Lane graph based on a Lanelet from the \ind{} dataset.]{\includegraphics[width=0.5\textwidth]{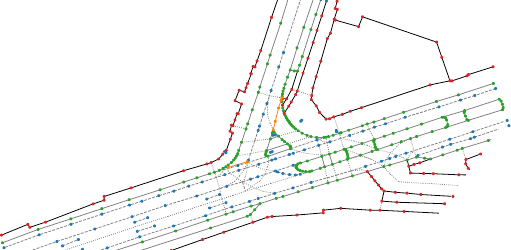}%
        \label{fig_second_case}}
        \hfill
    \subfloat[Lane graph based on a Lanelet from the \sind{} dataset.]{\includegraphics[width=0.5\textwidth]{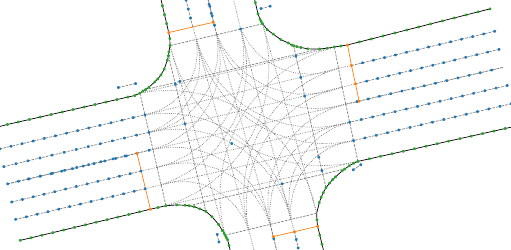}%
        \label{fig_third_case}}
        \hfill
    \subfloat[Lane graph based on a Lanelet from the \interact{} dataset.]{\includegraphics[width=0.5\textwidth]{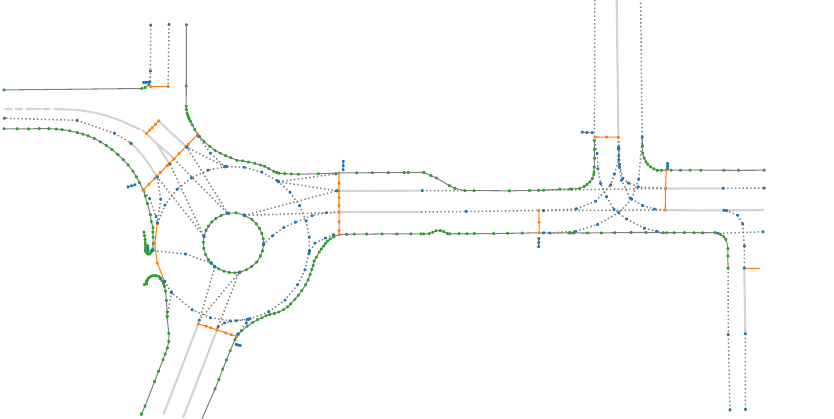}%
    \label{fig_fourth_case}}
    \caption{Example lane graphs for the \round{}, \ind{}, \sind{}, and \interact{} datasets.
    The lane graphs are constructed using the available Lanelet files, which contain semantic information about the scene.
    The lane graphs are constructed by extracting lane boundaries and markings, coloring them according to their type, and connecting them with edges.
    Although these examples are simple, combined with the available traffic data, they could be sufficient for many trajectory prediction tasks.
    Using the provided functionality, users can develop more complex lane graphs based on the desired level of detail.
    }
    \label{fig:lane-graph}
\end{figure*}

\subsection{High-Definition Maps}
\label{sec:lane-graph}
With the emergence of datasets that include map-based information, the effort to incorporate such contextual knowledge into the learning process has gained notable attention~\cite{liang2020learning,salzmann2020trajectron,deo2022multimodal,hu2022scenario,zhou2023query,westny2024diffusion}.
How to best represent this information in a way that is both expressive and computationally efficient remains an open research question.
One increasingly popular approach is to encode the semantics of the road network as a graph, with nodes representing geographical points of varying categories (traffic lights, intersections, lane boundaries, lane markings, buildings, \etc) and edges representing the connections between these points.
With the emerging interest in \glspl{GNN} for trajectory prediction tasks~\cite{huang2022survey,rahmani2023survey}, this representation has shown to be effective in capturing the spatial dependencies between agents and the environment while also ensuring consistent data modalities (as opposed to, \eg, raw images or rasterized maps).

All but the \highd{} dataset are accompanied by Lanelet~2~\cite{poggenhans2018lanelet2} files that can be used to extract semantic information about the scenes.
Our toolbox includes functionality for basic lane-graph construction, designed to serve as a foundation that users can modify according to their specific needs and preferences.
The resulting lane graph can be easily linked to agents in the scene, \eg, via a nearest-neighbor criterion. While the \highd{} dataset lacks Lanelet files, available lane marking information are used to construct a simplified lane graph. Examples are shown in Fig.~\ref{fig:lane-graph}.

\begin{lstlisting}[float, caption=Example batch of scenarios., label=lst:example_batch]
|\texttt{\color{codeblue}HeteroDataBatch}|(
    rec_id=[32],
    |\texttt{\color{codepurple}agent}|={
        num_nodes=350,
        ta_index=[32],
        atype=[350],
        inp_pos=[350, 15, 2],
        ...
        trg_pos=[350, 25, 2],
        ...
        input_mask=[350, 15],
        valid_mask=[350, 25],
        sa_mask=[350, 25],
        ma_mask=[350, 25],
        batch=[350],
        ptr=[33],
    },
    |\texttt{\color{codepurple}map\_point}|={
        num_nodes=14880,
        mtype=[14880],
        position=[14880, 2],
        batch=[14880],
        ptr=[33],
    },
    |\texttt{\color{codepurple}(map\_point, to, map\_point)}|={
        edge_index=[2, 31744],
        etype=[31744, 1],
    }
)
\end{lstlisting}

\subsection{Data Structure}
As scenarios contain a variable number of agents, the toolbox uses \gls{PYG}~\cite{fey2019pyg} when loading samples to handle this dimension inconsistency.
An example batch of scenarios is illustrated in Listing~\ref{lst:example_batch}.
The batch object contains two heterogeneous graphs, one for the agents and one for the lane graph.
The entries in the agent graph contain information on the agent features, masking tensors, and pointers to the batched data.
It is important to note that although \gls{PYG} components are included in our toolbox, this does not imply a requirement for users to incorporate \glspl{GNN} into their model, nor is it required for the preprocessing steps.
\gls{PYG} is mainly used for the convenience of handling the data.

\section{Evaluation Metrics}
\label{sec:evaluation}
This section outlines the most commonly used metrics from the literature to evaluate motion prediction models, which are incorporated in the toolbox.
The metrics are presented here for a single agent but can easily be extended to multi-agent problems by averaging over all scored agents in the scenario.
Here, $\state{k} \in \R^2$ refers to the ground truth position at time step $k$, $\stateestim{k}$ refers to the respective prediction, and $\predhrz$ is the prediction horizon steps.

For multimodal predictors, it is important to consider which of the prediction candidates should be used.
A commonly used approach is to evaluate using the prediction closest to the ground truth~\cite{huang2022survey,argoverse2dataset,ettinger2021large}.
While this approach overlooks mode probability, it facilitates comparison with models that do not explicitly parameterize modes, such as sampling-based generative models~\cite{zhao2019multi,ma2021continual,mao2023leapfrog,westny2024diffusion,choi2024dice}, where a \emph{best-of-many} approach is often applied instead. 
The toolbox includes various options to evaluate multimodal predictors, using either best-of-many or predicted mode probability.
For completeness, we also include metrics that consider predicted distributions.

\begin{itemize}%
	\item \emph{\gls{ADE}}: The average $L^2$-norm over the complete prediction horizon is
	\begin{equation}
		\label{eq:ade}
		\text{ADE} = \frac{1}{\predhrz}\sum_{k=1}^{\predhrz} \norm{\stateestim{k} - \state{k}}_2,
	\end{equation}
    where the minimum \gls{ADE} (min\gls{ADE}\textsubscript{$K$}) is commonly used to indicate the minimum value over $K$ predictions.
	\item \emph{\gls{FDE}}: 
	The $L^2$-norm of the final predicted position reflects the accuracy of the model in forecasting distant future events
	\begin{equation}
		\label{eq:fde}
		\text{FDE} = \norm{\stateestim{\predhrz} - \state{\predhrz}}_2,
	\end{equation}
    where the minimum \gls{FDE} (min\gls{FDE}\textsubscript{$K$}) is commonly used to indicate the minimum value over $K$ predictions.
    \item \emph{\gls{APDE}}: The average minimum $L^2$-norm between the predicted positions and the ground truth is used to estimate the path error.
	This can be used to determine predicted maneuver accuracy
	\begin{align}
		\label{eq:apde}
		\begin{split}
			\text{APDE} &= \frac{1}{\predhrz}\sum_{k=1}^{\predhrz} \norm{\stateestim{k} - \state{k^*}}_2 \\
			k^* &= \argmin_i \norm{\stateestim{k} - \state{i}}_2
		\end{split}
	\end{align}
	\item \emph{\gls{MR}}: The ratio of cases where the predicted final position is not within $2$~m of the ground truth is called the miss rate~\cite{huang2022survey}. Although the $2$~m threshold is derived from urban driving datasets, it has been shown to work well in highway scenarios as well~\cite{westny2023mtp,westny2024diffusion}.
	\item \emph{\gls{CR}}: The ratio of cases where for any timestep over the prediction horizon, the predicted position for another track was within $1$~m is called the collision rate. Note that this is multi-agent specific and may not apply to all studies.
\end{itemize}
\begin{itemize}%
	\item \emph{\gls{BFDE}}: 
	The \gls{BFDE} is a metric that considers the final position of the predicted distribution~\cite{argoverse2dataset}, which is particularly useful for multimodal predictors:
	\begin{equation}
		\label{eq:bfde}
		\text{FDE} = (1 - \mix{j})^2\norm{\stateestim{\predhrz}^j - \state{\predhrz}}_2,
	\end{equation}
    where $\mix{j}$ is the predicted mode probability of the $j$-th mode.
	\item \emph{\gls{ANLL}}: This metric provides an estimate of how well the predicted distribution matches the observed data:
	\begin{equation}
		\text{ANLL} = \frac{1}{\predhrz}\sum_{k=1}^{\predhrz} -\log \mathcal{D}( \state{k} | \stateestim{k}),
	\end{equation}
    where $\mathcal{D}$ is the predicted distribution, commonly modeled as a mixture of Gaussian or Laplacian distributions~\cite{deo2018convolutional,messaoud2020attention,westny2023mtp,zhou2023query}.
\end{itemize}
Minimum distance metrics are based on the mode with the lowest \gls{FDE} in the toolbox to ensure that the evaluations are consistent with the chosen prediction.

\section{Future Directions}
\label{sec:future-directions}
\begin{figure}[!t]
    \centering
    \includegraphics[width=0.9\columnwidth]{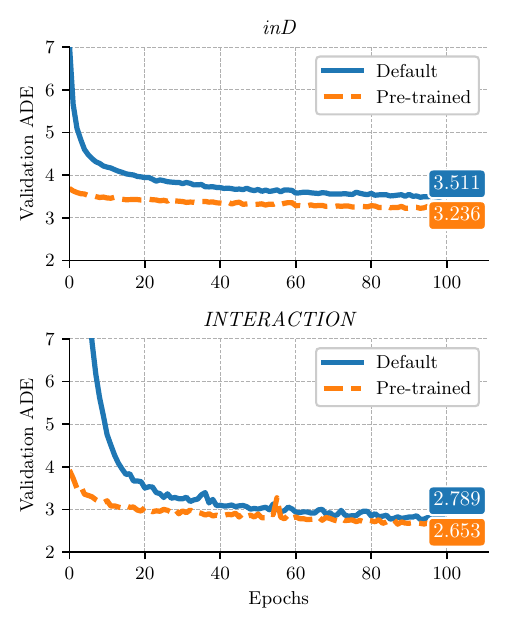}
    \caption{Impact of transfer learning on validation performance for the \ind{} and \emph{INTERACTION} datasets.
    Fine-tuning after pretraining on the other dataset yields lower initial error and improved performance throughout training.
    These results highlight the potential of transfer learning for improved generalization and motivate further research in this direction.}      
    \label{fig:transfer_learning}
\end{figure}

The introduction of the toolbox for the development of trajectory-prediction models for BEV datasets opens up several directions for future research.
Here we outline some potential areas for further exploration.
\begin{itemize}[leftmargin=*,label={\scriptsize\raisebox{0.1ex}{\ding{69}}}]
    \item \textbf{Scenario-Specific Performance:} Investigate and compare model performance across the dataset suite to determine whether a model structure or hyperparameter configuration adopted to one scenario retains its accuracy in another.
    \item \textbf{Zero-Shot Generalization:} Explore model generalization and zero-shot learning across datasets to investigate how models can be applied or adapted across different scenarios.
    \item \textbf{Transfer Learning:} Explore the potential of applying models, initially trained on one or more BEV datasets with subsequent fine-tuning, to other datasets within the suite or even to external datasets such as \emph{NuScenes}, \emph{Argoverse 2}, or \gls{WOMD}. An example of how transfer learning can be applied to the \emph{INTERACTION} and \ind{} datasets using a simple encoder-decoder model is shown in Fig.~\ref{fig:transfer_learning}.
\end{itemize}

\section{Conclusions}
\label{sec:conclusions}
This paper has presented a set of practices for developing trajectory prediction models using BEV datasets.
The proposals have been motivated by analyzing the most common practices in the field.
The methods have been implemented in an open-source toolbox
designed to standardize preprocessing and performance evaluation procedures---thereby simplifying comparative studies.
By alleviating the technical burden of developing research pipelines, the toolbox augments the potential for exploration and discoveries in the field of trajectory prediction.

\bibliographystyle{IEEEtran}
\bibliography{IEEEabrv,references.bib}{}
\end{document}